\documentclass[10pt,journal,compsoc]{IEEEtran}
\usepackage{graphicx}
\usepackage{color,soul}
\usepackage{hyperref}
\usepackage{amsmath}
\usepackage{float}

\ifCLASSOPTIONcompsoc
  \usepackage[nocompress]{cite}
\else
  \usepackage{cite}
\fi
\ifCLASSINFOpdf
\else
\fi
\begin{document}
%
\title{Bias Reduction via Cooperative Bargaining in Synthetic Graph Dataset Generation}
%
%
%
%

\author{Axel~Wassington and 
        Sergi~Abadal
        \thanks{A. Wassington and S. Abadal are with the Department of Computer Architecture, Universitat Polit\`{e}cnica de Catalunya, 08034 Barcelona, Spain.}
        }

\IEEEtitleabstractindextext{%
\begin{abstract}
In general, to draw robust conclusions from a dataset, all the analyzed population must be represented on said dataset. Having a dataset that does not fulfill this condition normally leads to selection bias. Additionally, graphs have been used to model a wide variety of problems. Although synthetic graphs can be used to augment available real graph datasets to overcome selection bias, the generation of unbiased synthetic datasets is complex with current tools. In this work, we propose a method to find a synthetic graph dataset that has an even representation of graphs with different metrics. The resulting dataset can then be used, among others, for benchmarking graph processing techniques as the accuracy of different Graph Neural Network (GNN) models or the speedups obtained by different graph processing acceleration frameworks.
\end{abstract}

}

\maketitle

\IEEEdisplaynontitleabstractindextext

%
\IEEEpeerreviewmaketitle

\IEEEraisesectionheading{\section{Introduction}\label{sec:intro}}

%
%
%
%
\IEEEPARstart{S}{ystems from} different areas of knowledge can be modeled as graphs; from social science to technological networks, among many others \cite{newman2003structure}. Graphs from different areas can have different characteristics. For example, the clustering coefficient (also known as transitivity) in a power grid may be low, because its objective is to cover vast areas, and that makes graphs representing power grids more tree-like. On the other hand, the clustering on social relationship graphs may be high because one normally gets to know the friends of friends, so the friendship relationship has some level of transitivity that is reflected as a high clustering coefficient on the graph representation.

When benchmarking graph processing techniques, for example analyzing the accuracy of Graph Neural Networks (GNN) models \cite{scarselli2008graph} or the speedup obtained by GNN accelerators \cite{abadal2021computing}, it is common to use a limited set of graphs. In some cases, a standardized dataset exists to be able to compare different approaches. In the case of GNNs, the Open Graph Benchmark (OGB) suite \cite{hu2020ogb} has been recently proposed to that end. The problem is that these datasets can suffer from selection bias. This means that some areas of the graph space may stay unexplored or underrepresented. OGB, for example, does not contain graphs that represent road networks. Hence, solutions that work for the graphs on OGB, may not be good for logistics problems that involve road networks, because such networks have a set of characteristics that are underrepresented in the OGB datasets.

In general, to draw robust conclusions from a dataset, all the analyzed population must be represented on said dataset, or, in other words, the dataset must explore the solution space relevant to the problem as comprehensively as possible. This is in particular important for machine learning, where an underexplored solution space can lead to biased results \cite{cortes2008sample}. In some areas, a common approach to overcome this bias is to augment the dataset with synthetic data. For example, on \cite{kortylewski2019analyzing}, synthetically generated images are used to improve face recognition, whereas on \cite{jaipuria2020deflating}, simulation scenarios are used to train autonomous driving in different conditions. Also, on \cite{draghi2021bayesboost}, synthetic data is generated based on real patient data to train AI models. Similarly, we intend to use synthetic graphs to augment the available datasets.

\begin{figure}
    \centering
    \includegraphics[width=\columnwidth]{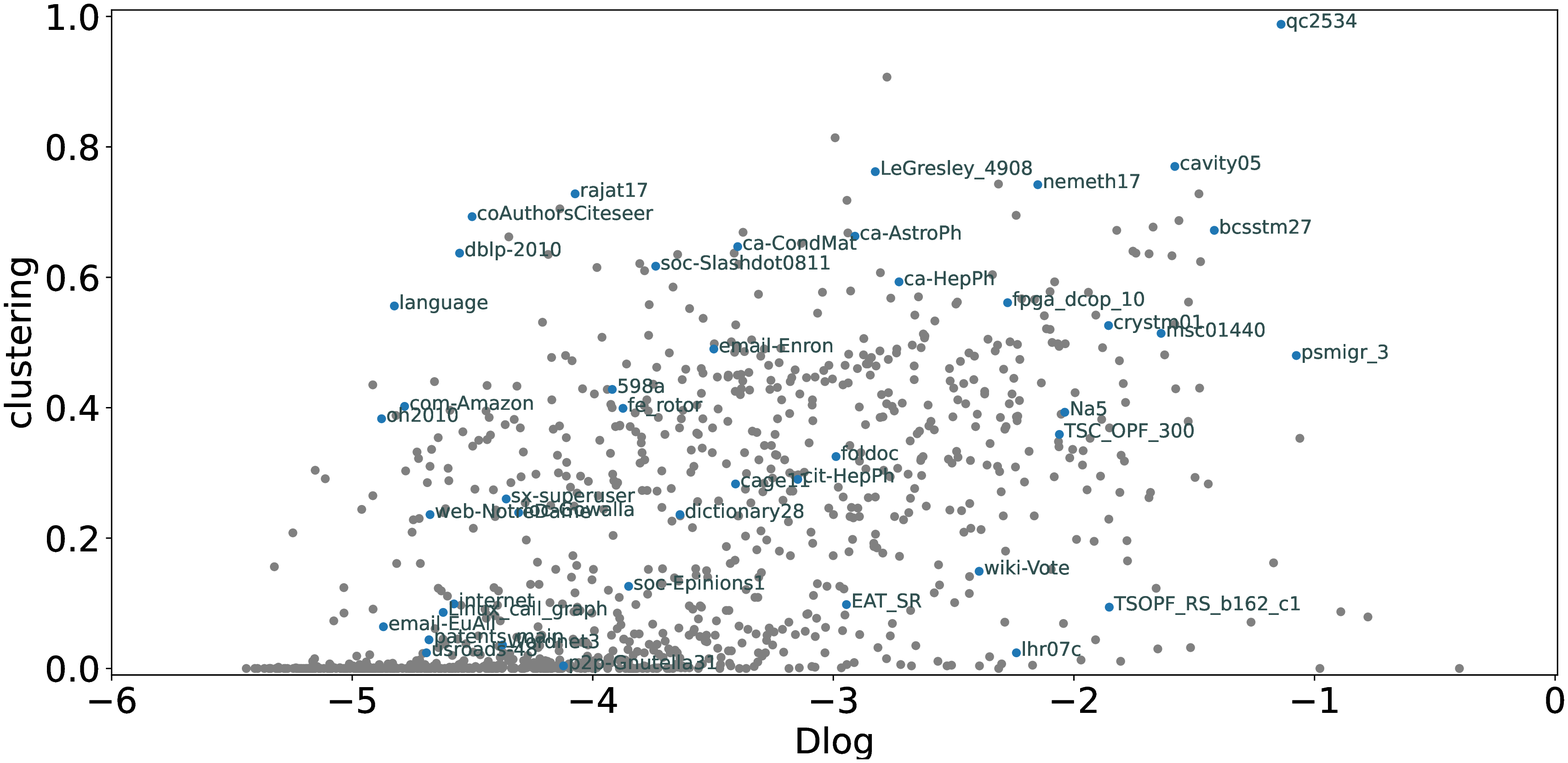}
    \vspace{-0.6cm}
    \caption{A sample of the naive Baseline dataset (in gray) compared to the Validation dataset of real graphs (in blue) on a projection over the analyzed metrics.}
    \label{fig:baseline_validation}
    \vspace{-0.2cm}
\end{figure}

We can think of the problem of synthetic graph dataset generation as doing a sample of the space of all graphs. As such, the risk of introducing bias by sampling more graphs with certain characteristics is still present. As an example, the use of a naive graph generation strategy is shown in Figure \ref{fig:baseline_validation}. This naive method (also referred to as the baseline method later in the paper) clearly leads to a dataset with areas in the metric space that are underrepresented, but that are common on real graphs.

\begin{figure*}[!t]
    \centering
    \includegraphics[width=0.95\textwidth]{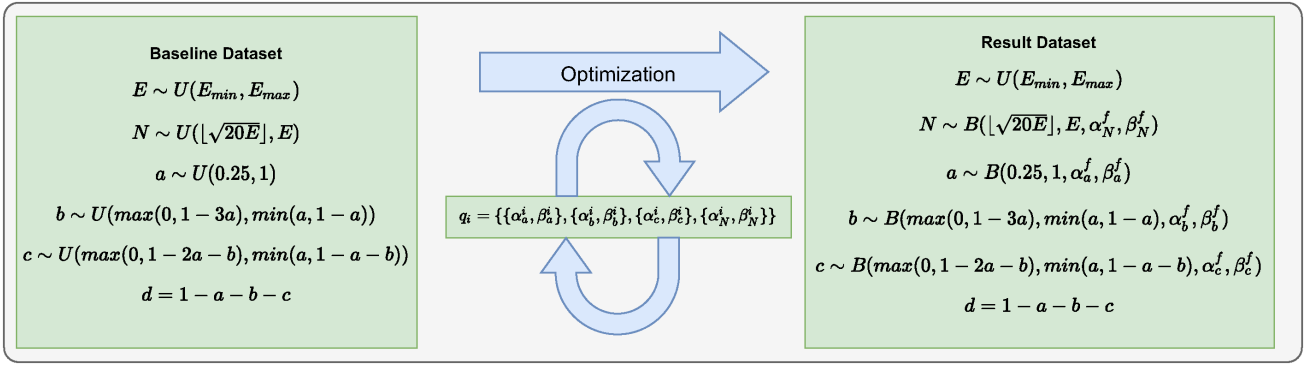} \vspace{-0.2cm}
        \caption{An overview of the optimization process, where on the left we can see the distributions for the baseline dataset, on the middle the parameters that are optimized, and on the right the distributions for the result dataset}
        \label{fig:opti}\vspace{-0.2cm}
\end{figure*}

Efforts to mimic real graphs or datasets using synthetic methods have been widely studied. For example, RMAT proposes a method to mimic a real graph by using a specific set of parameters derived from the graph characteristics in \cite{chakrabarti2004r}. Others try to mimic the degree sequence as the configuration model (analyzed in \cite{newman2003structure}) and the expected degree model \cite{chung2002connected}. More sophisticated methods involve using optimization techniques for the exploration of graphs with specific characteristics as in  \cite{10.1007/978-3-319-58943-5_45}, where graphs are generated using evolutionary computing. In contrast, to the best of our knowledge, the generation of a graph dataset that covers the metric space evenly has not been explored. 

To bridge this gap, the objective of this study is to generate a dataset of graphs that explores the solution space for certain graph metrics. To meet this objective, we will take base on the popular RMAT random graph generator \cite{chakrabarti2004r}, which can produce graphs from close to power-law degree distribution to one with binomial degree distribution (as in the Erdös-Rényi random graph model). The variety of graphs that this generator can produce is important for the objective of the study. A mathematical analysis of RMAT generated graphs is made on \cite{https://doi.org/10.1002/net.20417}, where the variety of degree distribution is explained in more detail. Still, using RMAT in a naive way leads to the results shown in Figure \ref{fig:baseline_validation}, where most real graphs are underrepresented. This is because most combinations of RMAT parameters generate graphs with low clustering coefficient, and only some specific combinations generate graphs with higher clustering coefficient, as can be seen in Figure \ref{fig:param_clustering_original}.

In this work, we propose a method to find a distribution of the RMAT parameters that will generate the most evenly distributed graph dataset in terms of a series of metrics. We present a method based on an optimization process summarized in Figure \ref{fig:opti}, where the distribution of the parameters of the graph generator are adjusted to achieve an evenly distributed result dataset. To demonstrate the effectiveness of the approach, we first present a naive solution and compare it with real validation graphs dataset from the SparseMatrix collection \cite{Kolodziej2019}. Then, we compare the resulting dataset with the same validation dataset to show how the final dataset covers the space of the real graphs and has a greater variety in terms of the selected metrics. On top of that, we make available an open-source tool, called Graphlaxy \footnote{Available at \url{https://github.com/BNN-UPC/graphlaxy}}, which allows to replicate the study and generate evenly distributed datasets. This can be of great relevance in graph processing or GNN problems whose popularity has increased greatly in the recent years \cite{garg2022understanding}.

The rest of this paper is organized as follows. In Section \ref{sec:motiv}, we analyze the baseline dataset, which will be used as starting point of the optimization. In Section \ref{sec:method}, we provide a mathematical description of the optimization problem used to find the result dataset, that is the evenly distributed dataset that we obtain from the method. In Section \ref{sec:results}, we discuss the parameters used for the optimization and the results. Finally, in Section \ref{sec:conclusions}, we conclude the article.

\section{Motivation}
\label{sec:motiv}

In this section, to understand why this process is needed, we will naively create a baseline dataset and show that this dataset is biased. This baseline dataset will then be used to compare the results, and as data to approximate the objective function, discussed in Section \ref{sec:method}. To understand how such a baseline dataset is created, we first summarize the operation of RMAT. 

The RMAT generator uses six parameters. The first two parameters are the number of nodes $N$ and the number of edges $E$. Then, we have a vector $r = [a,b,c,d]$ that defines a probability distribution of an edge for each pair of nodes, similar to the fitness model proposed in \cite{bianconi2001competition}, but considering pairs of nodes instead of attachment to each node. The $r$ vector is a probability vector and, as such, the sum of its elements must be 1. The RMAT generator works by dividing the adjacency matrix into four quadrants recursively and assigning the probability of an edge falling in each of the groups following the $r$ vector.

The baseline dataset is generated in a naive way, using uniform distributions for all RMAT parameters. In particular, the parameter distribution shown on the left side of Figure \ref{fig:opti} was used, where $U(a,b)$ stands for the uniform distribution in the interval $[a,b]$. The lower-bound for $N$ is a rough approximation of the relation between $N$ and $E$, to form a graph with a density of $0.1$, which is the maximum density that was found on real graphs and is derived from the equation $\frac{2E}{N(N-1)} \leq \frac{1}{10}$. Similarly, the upper bound is the maximum value that $N$ can take to make a connected graph. The limits of the parameters of the $r$ vector were set so that $a \geq b$, $a \geq c$, and $a \geq d$, as recommended by the RMAT study.

\begin{figure}[!t]
\centering
    \includegraphics[width=\columnwidth]{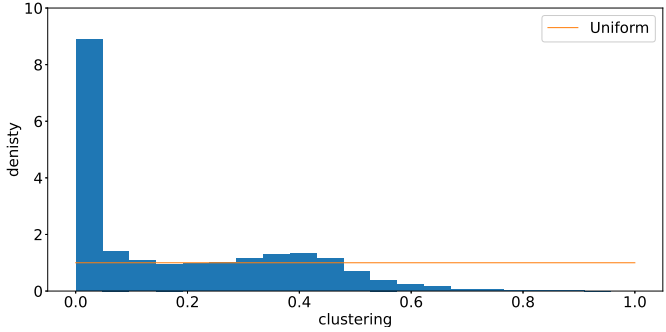}
    \vspace{-0.6cm}
        \caption{Clustering coefficient density on the Baseline dataset.}
        \label{fig:clustering_original}
        \vspace{-0.2cm}
\end{figure}

\begin{figure}[!t]
 \centering
    \includegraphics[width=\columnwidth]{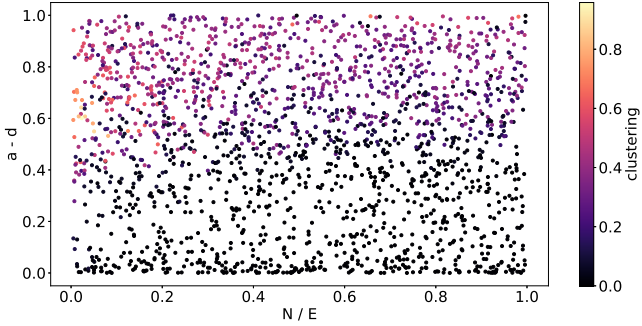}
    \vspace{-0.6cm}
    \caption{Impact of the RMAT parameters on the clustering of the generated graphs. The x-axis corresponds to parameters $N$ and $E$ ($N$ and $E$ are not the final numbers of edges and nodes, but the model parameters). The y-axis corresponds to the difference between the $a$ and $d$ parameters. The colors indicate the mean clustering coefficient of each graph.}
    \label{fig:param_clustering_original}
    \vspace{-0.2cm}
\end{figure}

Figure \ref{fig:clustering_original} illustrates that the baseline dataset is biased towards graphs with low clustering coefficients. This is one of the motivations to generate a more diverse dataset. Figure \ref{fig:param_clustering_original} shows how the clustering coefficient is affected by the parameters of the RMAT model. The plot presents how the relation between $N$ and $E$ parameters, and the difference between $d$ and $a$ parameters have a relationship with the clustering coefficient of the resulting graphs.

Based on the above, we can conclude that using this naive method of generating graph datasets is biased, leaving the high clustering graphs underrepresented and that we need to favor certain parameter combinations to get a less biased dataset.

\section{The Proposed Method}
\label{sec:method}
In this section, we describe the method used to find an evenly distributed dataset. We will start with the construction of the function to optimize, then we will present the cooperative bargaining optimization, and finally some details of how the optimization was applied to RMat parameters and the metrics we selected for this study. The general objective of the methodology is to find a distribution for the parameters of the graph generator (RMAT) to generate a result dataset with an even distribution on the metric projection. An overview of the process can be seen in Figure \ref{fig:opti}, where the optimization process is used to find the parameters of the distribution used in RMAT to generate the result dataset.

This method can be generalized into weighted, directed, and non-connected graphs, but for simplicity, in this study, we will use undirected, unweighted, and connected graphs. For the generalization, it is necessary to rethink the definition of the metrics so that they make sense in each scenario. Since RMAT produces disconnected and directed graphs, we transform the graphs generated by RMAT into undirected graphs and keep the biggest connected component before starting the analysis.

\subsection{The probability distribution}
\label{ssec:of}

\begin{figure*}
    \centering
    
    \includegraphics[width=0.8\textwidth]{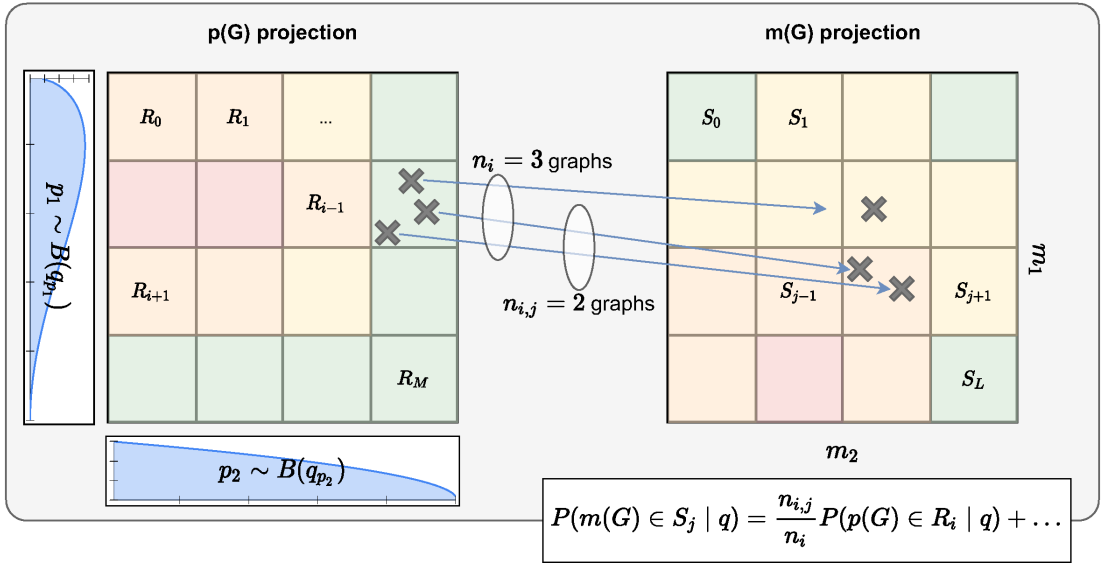}
    \vspace{-0.4cm}
        \caption{An overview of the concept behind the building of the function to optimize. It represents two projections of the graph, divided into grids to discretize the problem. The left projection represents the parameters used to build the graph, and the right projection the metrics of the graph. The colors of the cells represent how probable is for a graph to fall into that cell in the projection.}
        \label{fig:prob}
        \vspace{-0.2cm}
\end{figure*}

In Figure \ref{fig:prob} we present the idea behind the construction of the function to optimize, where the conditional probability of a graph having a particular combination of metrics is approximated using the data on the baseline dataset. This subsection will go through the construction of this mathematical expression and its meaning.

To understand the relationship between the parameters and the metrics of a graph we define two projections of a graph. The metric projection $m(G) = [m_1, m_2, \dots, m_n]$ is the vector of all the relevant metrics of the graph $G$, and the parameter projection $p(G) = [p_1, p_2, \dots, p_l]$ is the vector of the graph generator parameters used to generate the graph $G$. Each of these projections defines a vector space that we call metric space and parameter space respectively.

Then the objective function can be defined as finding an evenly distributed projection of the graph on the metric space. Or in other words, we want the probability of a graph taking any given combination of metrics to be constant. 

To discretize the problem we define a grid on each of the spaces. The metric space grid is $S = \{S_1, S_2, \dots , S_M\}$ where each cell on the grid can be defined as $S_i=[s^1_{i1}, s^2_{i1}) \times [s^1_{i2}, s^2_{i2}) \times \dots \times [s^1_{iM}, s^2_{iM})$, with $s^1_{ij}$ being the minimum and $s^2_{ij}$ the maximum of cell $j$ on dimension $i$, and with each dimension representing a metric. Analogously, the parameter space grid is $R = \{R_1, R_2, \dots , R_L\}$, where each cell on the grid can be defined as $R_i=[r^1_{i1}, r^2_{i1}) \times [r^1_{i2}, r^2_{i2}) \times \dots \times [r^1_{iL}, r^2_{iL})$, with $r^1_{ij}$ being the minimum and $r^2_{ij}$ the maximum of cell $j$ on dimension $i$, and with each dimension representing a parameter of the graph generator.

With this definition, we can say that the objective of the optimization is to find a parameter distribution with parameters $q$ that makes the probability of the graph metric projection fall in each of the cells of the grid constant,
\begin{equation}
    P(m(G) \in S_j | q) = \frac{1}{M}, \forall j \in [0,L].
\end{equation}

We can now, using the law of total probability, find an expression of this probability in terms of the probability of falling on a cell in the grid of the parameter projection as in
\begin{multline}
    P(m(G) \in S_j | q) = \\
    \sum_{i = 0}^{M} 
    P( m(G) \in S_j \mid p(G) \in R_i)\times
    P(p(G) \in R_i | q).
\label{eq:total}
\end{multline}

The probability of a graph $G$ falling in $S_i$ given that it has fallen in $R_j$ can be approximated by using the baseline dataset as follows,
\begin{equation}
    P( m(G) \in S_j \mid
  p(G) \in R_i) \approx \frac{n_{i,j}}{n_i},
\end{equation}
where $n_i$ is the number of graphs on the baseline dataset whose parameter projection fall into cell $R_i$, and $n_{i,j}$ is the number of graphs on the baseline dataset whose parameter projection fall in $R_i$ and whose metric projection fall in cell $S_j$. This is the percentage of graphs from $R_i$ that also fall on $S_j$.

The other part of Equation \eqref{eq:total} can be obtained from the probability function that will be used to generate the parameters
\begin{equation}
    P( p(G) \in R_i | q) =\prod_{k =0}^l   F(r^2_{ik}; q_k) - F(r^1_{ik}; q_k)),
\end{equation}
where F(a;q) is the cumulative probability function evaluated on $a$ with parameters $q$.

This means that we can now calculate the resulting discrete distribution on the metric projection, given that we know the parameters $q = {q_1, q_2, \dots}$ that will be used to generate the result dataset.

\subsection{The optimization}
We define an optimization problem to find the evenest distribution on the graph's metric projections. The variables to optimize are the parameters of the distributions of the graph generator parameters $q^i= \{q^i_1, q^i_2, \dots\}$, where $q^i_k$ is the parameter list for the distribution of the graph generator parameter $k$ on the optimization generation $i$.

With the probability distribution we defined on the previous subsection, together with the concept of Nash barganing schema \cite{nash1950bargaining}, the objective function is defined as
\begin{equation}
    f(q) = - \frac{1}{M}\sum_{i=1}^M \log_2\left(1 + (M-1)  P(m(G) \in S_i \mid q)\right).
\end{equation}
This equation can be thought of as cooperative bargaining between the metric grid cells to get the highest probability of a graph falling on them.

On the one hand, the function $f(q)$ reaches its maximum when the probability is concentrated on only one cell, at $f_{max} = \tfrac{-log_2(M)}{M}$, which gets closer to 0 the more cells the grid has. On the other hand, the minimum of $f(q)$ is found when the probability is evenly distributed through all the cells at $f_{min} = -log_2(2 - \tfrac{1}{M})$, a value that gets closer to -1 the more cells the grid has. This means that the lower the value of the function is, the more even the resulting distribution will be.

\subsection{Application of the method}

In this study, we used the Beta distribution with location and scale for the distribution of each of the parameters \cite{olea2011use}. The Beta distribution has the advantage of being able to adapt to different shapes depending on its parameters, at the same time that it only uses two parameters (not counting the location and scale). The parameters of the distribution are $\alpha$ and $\beta$.

The RMAT parameter distributions to optimize are those of $N$, $a$, $b$ and $c$. This is because we use a uniform distribution for $E$, and $d$ can be calculated from the other parameters of the vector $r$. The formulation of the distribution of each of these parameters can be seen on the right part of Figure \ref{fig:opti}, where $B(s,t,\alpha,\beta)$ stands for the Beta distribution displaced and scaled to be defined in the interval $[s,t]$ with parameters $\alpha$ and $\beta$ and $f$ is the final step of the optimization. This leave us with a vector $q^k = \{\{\alpha_N^k, \beta_N^k\}, \{\alpha_a^k, \beta_a^k\}, \{\alpha_b^k, \beta_b^k\}\}$ of parameters to optimize. The initial value for the $\alpha$ and $\beta$ for each of the distributions was set to 1 (no information) and was bounded to the interval $(0,100]$.

With all this, the method has as parameters: the number of graphs on the baseline dataset $n$, the edge interval for the baseline dataset $[E_{min}, E_{max}]$, the number of cells on the metric space $M$ and the number of cells on the parameter space $L$.

\section{Results}
\label{sec:results}
The parameters used to illustrate the method are $n = 10000$ graphs on the Baseline dataset with a number of edges in the interval $[E_{min},E_{max}] = [10^{5}, 10^{6}]$, a metric grid of $10\times10$ ($M=100$) cells, and a parameter grid of $20^4$ ($L=160000$) cells, but most of this cells are empty because the possible values of $b$ and $c$ depend on the value of $a$, with only $4403$ cells with at least one data point. While the method is generalizable to any graph characterization metric, in this work we illustrate it using the clustering coefficient and the $log_{10}$ of the density of the graph as metrics.

An analysis, using different samples of the Baseline dataset, was made to determine the best number of generations to run the optimization and avoid overfitting. The results shown in this section are after selecting an appropriate tolerance to avoid overfitting.

 \begin{figure}
    \centering
    \includegraphics[width=\columnwidth]{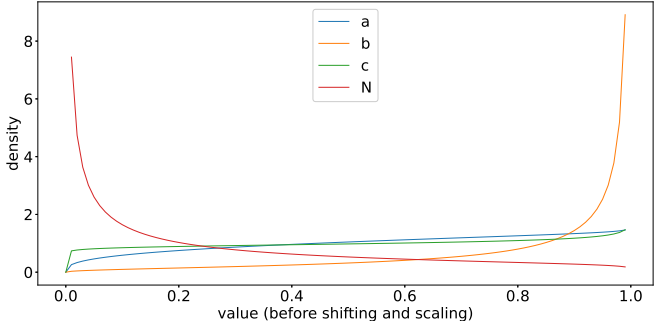}
    \vspace{-0.6cm}
    \caption{The resulting Beta distribution for each of the RMAT parameters.}
    \label{fig:beta}
    \vspace{-0.2cm}
\end{figure}

 \begin{figure}
    \centering
    \includegraphics[width=\columnwidth]{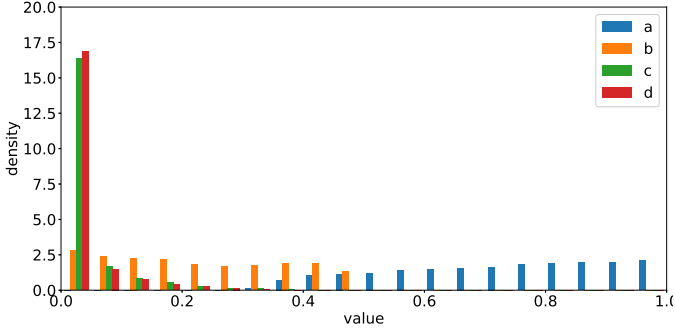}
    \vspace{-0.6cm}
    \caption{Density of the RMAT parameters after shifting and rescaling.}
    \label{fig:param_density}
    \vspace{-0.2cm}
\end{figure}

After applying our method, the resulting parameters are $\alpha_a^f = 1.35$, $\beta_a^f = 0.98$, $\alpha_b^f = 1.46$, $\beta_b^f = 0.23$, $\alpha_c^f = 1.06$, $\beta_c^f = 0.91$, $ \alpha_N^f = 0.35$ and $\beta_N^f = 1.16$, with a fit of $f(S) = -0.80$ in 7 generations. This is an improvement with respect to the Baseline dataset with a fit $f(I) = -0.65$. With respect to the meaning of the resulting parameters, we can see on Figure \ref{fig:beta} the shape of the resulting beta distributions, where we can see that the method favors small values for parameter $N$, big values for parameter $b$, and has a small preference for big values for parameters $a$ and $c$. The resulting distribution of parameters can be seen on Figure \ref{fig:param_density}, where se can see that parameters $a$ and $b$ are balanced, but parameters $c$ and $d$ have a higher density of small values, the preference of low values of $N$ and $d$, and higher values of $a$ was expected from the information given by Figure \ref{fig:param_clustering_original}, where the zone with higher clustering has these characteristics.

In Table \ref{tab:stats}, we can see how the correlation and covariance between the metrics of the Result dataset are maintained low. Some correlation is expected between both metrics, because the denser a graph is, the higher the clustering coefficient can be, being the extreme cases the complete graph with clustering of 1, and a density of 1, and a tree with a clustering of 0 and the minimum possible density for the number of nodes. A high value on correlation or covariance could indicate that some combination of both variables is kept unexplored.

 \begin{figure}
    \centering
    \includegraphics[width=\columnwidth]{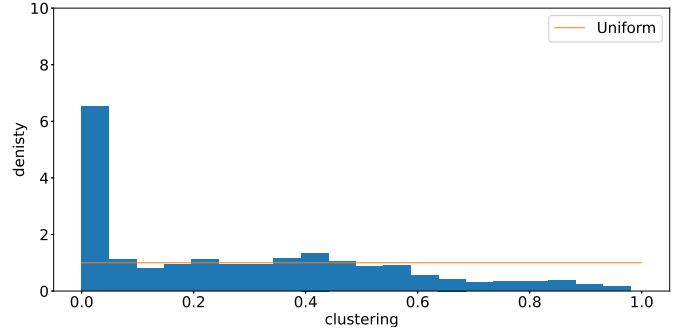}
    \vspace{-0.6cm}
    \caption{Clustering coefficient density on the Result dataset.}
    \label{fig:clustering_density_res}
    \vspace{-0.2cm}
\end{figure}

 \begin{figure}
    \centering
    \includegraphics[width=\columnwidth]{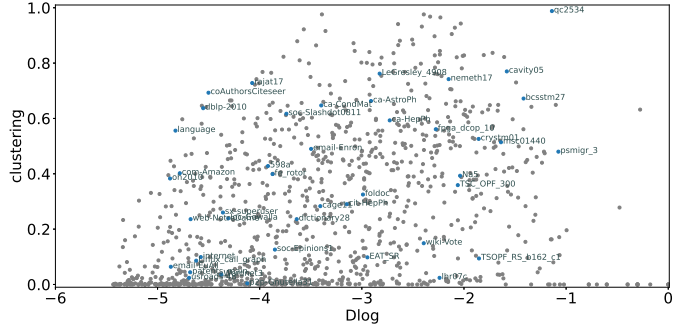}
    \vspace{-0.6cm}
    \caption{Sample of the Result dataset (in gray) and the Validation dataset of real graphs (in blue) on a projection over the analyzed metrics.}
    \label{fig:validation}
    \vspace{-0.2cm}
\end{figure}

We can see by comparing Figure \ref{fig:clustering_density_res} and Figure \ref{fig:clustering_original}, that the clustering of the result dataset is distributed more evenly in the Result dataset, which was one of the motivations of the study. The same applies, though more subtle, to the density.

To validate the method, a set of 42 graphs from SparseSuite \cite{Kolodziej2019} with the number of edges between 10000 and 150000 was used. In Figure \ref{fig:validation}, it can be seen that the range of the assessed parameters of the real graphs is covered in most cases by the generated dataset cloud of points.

\begin{table}
\centering

\caption{Correlation and covariance between clustering and $Dlog = log_{10}(density)$, minimum and maximum of clustering ($C$) and $Dlog$ for each of the datasets.} 
\label{tab:stats}
\begin{tabular}{l|r|r|r|r|r|r}
Name       & \multicolumn{1}{l|}{Corr} & \multicolumn{1}{l|}{Cov} & \multicolumn{1}{l|}{$C_{min}$} & \multicolumn{1}{l|}{$C_{max}$} & \multicolumn{1}{l|}{$Dlog_{min}$} & \multicolumn{1}{l}{$Dlog_{max}$} \\ \hline
Validation & 0,15                      & 0,02                     & 0,01                        & 0,67                        & -4,87                          & -1,89                         \\
Baseline   & 0,60                      & 0,11                     & 0,00                        & 0,89                        & -5,48                          & -0,99                         \\
Result     & 0,32                      & 0,09                     & 0,00                        & 0,98                        & -5,60                           & 0.00                         
\end{tabular}
\end{table}

\section{Conclusions}
\label{sec:conclusions}
In this study, we have analyzed the importance of unbiased datasets for robust conclusions and in particular for machine learning. Then, we have generated and analyzed a naive synthetic graph dataset and discovered why it was biased. Based on these discoveries, we have proposed an optimization problem using the concept of cooperative bargaining to avoid this bias. Finally, we validated the resulting dataset against a set of real graphs, showing that most of the graphs are represented on the synthetic dataset. Furthermore, we developed a command line tool using the described method which is now publicly available at \url{https://github.com/BNN-UPC/graphlaxy}. 

It is added value that the method defined in this study can be applied to other random graph generators and other graph metrics. 
The resulting dataset can benefit different areas of knowledge: in particular, this study was developed for benchmarking graph processing techniques, especially in the field of GNN, where a great variety of models and accelerators exist \cite{abadal2021computing}


\ifCLASSOPTIONcompsoc
  \section*{Acknowledgments}
\else
  \section*{Acknowledgment}
\fi

Authors gratefully acknowledge support from NEC Laboratories Europe GmbH through a Postdoctoral Fellowship. Authors would also like to thank Ramon Ferrer-i-Cancho and Daniel Thuerck for the discussions that helped in the development of the study.

\ifCLASSOPTIONcaptionsoff
  \newpage
\fi



\bibliographystyle{IEEEtran}
\bibliography{IEEEabrv, main}

\begin{thebibliography}{10}
\providecommand{\url}[1]{#1}
\csname url@samestyle\endcsname
\providecommand{\newblock}{\relax}
\providecommand{\bibinfo}[2]{#2}
\providecommand{\BIBentrySTDinterwordspacing}{\spaceskip=0pt\relax}
\providecommand{\BIBentryALTinterwordstretchfactor}{4}
\providecommand{\BIBentryALTinterwordspacing}{\spaceskip=\fontdimen2\font plus
\BIBentryALTinterwordstretchfactor\fontdimen3\font minus
  \fontdimen4\font\relax}
\providecommand{\BIBforeignlanguage}[2]{{%
\expandafter\ifx\csname l@#1\endcsname\relax
\typeout{** WARNING: IEEEtran.bst: No hyphenation pattern has been}%
\typeout{** loaded for the language `#1'. Using the pattern for}%
\typeout{** the default language instead.}%
\else
\language=\csname l@#1\endcsname
\fi
#2}}
\providecommand{\BIBdecl}{\relax}
\BIBdecl

\bibitem{newman2003structure}
M.~E. Newman, ``The structure and function of complex networks,'' \emph{SIAM
  review}, vol.~45, no.~2, pp. 167--256, 2003.

\bibitem{scarselli2008graph}
F.~Scarselli, M.~Gori, A.~C. Tsoi, M.~Hagenbuchner, and G.~Monfardini, ``The
  graph neural network model,'' \emph{IEEE transactions on neural networks},
  vol.~20, no.~1, pp. 61--80, 2008.

\bibitem{abadal2021computing}
S.~Abadal, A.~Jain, R.~Guirado, J.~L{\'o}pez-Alonso, and E.~Alarc{\'o}n,
  ``Computing graph neural networks: A survey from algorithms to
  accelerators,'' \emph{ACM Computing Surveys (CSUR)}, vol.~54, no.~9, pp.
  1--38, 2021.

\bibitem{hu2020ogb}
W.~Hu, M.~Fey, M.~Zitnik, Y.~Dong, H.~Ren, B.~Liu, M.~Catasta, and J.~Leskovec,
  ``Open graph benchmark: Datasets for machine learning on graphs,''
  \emph{arXiv preprint arXiv:2005.00687}, 2020.

\bibitem{cortes2008sample}
C.~Cortes, M.~Mohri, M.~Riley, and A.~Rostamizadeh, ``Sample selection bias
  correction theory,'' in \emph{International conference on algorithmic
  learning theory}.\hskip 1em plus 0.5em minus 0.4em\relax Springer, 2008, pp.
  38--53.

\bibitem{kortylewski2019analyzing}
A.~Kortylewski, B.~Egger, A.~Schneider, T.~Gerig, A.~Morel-Forster, and
  T.~Vetter, ``Analyzing and reducing the damage of dataset bias to face
  recognition with synthetic data,'' in \emph{Proceedings of the IEEE/CVF
  Conference on Computer Vision and Pattern Recognition Workshops}, 2019.

\bibitem{jaipuria2020deflating}
N.~Jaipuria, X.~Zhang, R.~Bhasin, M.~Arafa, P.~Chakravarty, S.~Shrivastava,
  S.~Manglani, and V.~N. Murali, ``Deflating dataset bias using synthetic data
  augmentation,'' in \emph{Proceedings of the IEEE/CVF Conference on Computer
  Vision and Pattern Recognition Workshops}, 2020, pp. 772--773.

\bibitem{draghi2021bayesboost}
B.~Draghi, Z.~Wang, P.~Myles, and A.~Tucker, ``Bayesboost: Identifying and
  handling bias using synthetic data generators,'' in \emph{Third International
  Workshop on Learning with Imbalanced Domains: Theory and Applications}.\hskip
  1em plus 0.5em minus 0.4em\relax PMLR, 2021, pp. 49--62.

\bibitem{chakrabarti2004r}
D.~Chakrabarti, Y.~Zhan, and C.~Faloutsos, ``R-mat: A recursive model for graph
  mining,'' in \emph{Proceedings of the 2004 SIAM International Conference on
  Data Mining}.\hskip 1em plus 0.5em minus 0.4em\relax SIAM, 2004, pp.
  442--446.

\bibitem{chung2002connected}
F.~Chung and L.~Lu, ``Connected components in random graphs with given expected
  degree sequences,'' \emph{Annals of combinatorics}, vol.~6, no.~2, pp.
  125--145, 2002.

\bibitem{10.1007/978-3-319-58943-5_45}
M.~Verstraaten, A.~L. Varbanescu, and C.~de~Laat, ``Synthetic graph generation
  for systematic exploration of graph structural properties,'' in
  \emph{Euro-Par 2016: Parallel Processing Workshops}, F.~Desprez, P.-F. Dutot,
  C.~Kaklamanis, L.~Marchal, K.~Molitorisz, L.~Ricci, V.~Scarano, M.~A.
  Vega-Rodr{\'i}guez, A.~L. Varbanescu, S.~Hunold, S.~L. Scott, S.~Lankes, and
  J.~Weidendorfer, Eds.\hskip 1em plus 0.5em minus 0.4em\relax Cham: Springer
  International Publishing, 2017, pp. 557--570.

\bibitem{https://doi.org/10.1002/net.20417}
\BIBentryALTinterwordspacing
C.~Groër, B.~D. Sullivan, and S.~Poole, ``A mathematical analysis of the r-mat
  random graph generator,'' \emph{Networks}, vol.~58, no.~3, pp. 159--170,
  2011. [Online]. Available:
  \url{https://onlinelibrary.wiley.com/doi/abs/10.1002/net.20417}
\BIBentrySTDinterwordspacing

\bibitem{Kolodziej2019}
\BIBentryALTinterwordspacing
S.~P. Kolodziej, M.~Aznaveh, M.~Bullock, J.~David, T.~A. Davis, M.~Henderson,
  Y.~Hu, and R.~Sandstrom, ``The suitesparse matrix collection website
  interface,'' \emph{Journal of Open Source Software}, vol.~4, no.~35, p. 1244,
  2019. [Online]. Available: \url{https://doi.org/10.21105/joss.01244}
\BIBentrySTDinterwordspacing

\bibitem{garg2022understanding}
R.~Garg, E.~Qin, F.~Mu{\~n}oz-Mart{\'\i}nez, R.~Guirado, A.~Jain, S.~Abadal,
  J.~L. Abell{\'a}n, M.~E. Acacio, E.~Alarc{\'o}n, S.~Rajamanickam
  \emph{et~al.}, ``Understanding the design space of sparse/dense multiphase
  dataflows for mapping graph neural networks on spatial accelerators,''
  \emph{Proceedings of the IPDPS'22}, 2022.

\bibitem{bianconi2001competition}
G.~Bianconi and A.-L. Barab{\'a}si, ``Competition and multiscaling in evolving
  networks,'' \emph{EPL (Europhysics Letters)}, vol.~54, no.~4, p. 436, 2001.

\bibitem{nash1950bargaining}
J.~F. Nash~Jr, ``The bargaining problem,'' \emph{Econometrica: Journal of the
  econometric society}, pp. 155--162, 1950.

\bibitem{olea2011use}
R.~A. Olea, ``On the use of the beta distribution in probabilistic resource
  assessments,'' \emph{Natural resources research}, vol.~20, no.~4, pp.
  377--388, 2011.

\end{thebibliography}
\onecolumn

\newpage
\appendix{Validation set}
\begin{table}[H]
\centering

\caption{Table describing the validation set, where $N$ is the number of nodes, $E$ the number of edges, $C$ is the mean clustering coefficient and $Dlog$ refers to the logarithm in base 10 of the graph density.}
\label{tab:val}
\begin{tabular}{l|l|l|l|l|l|l}
\textbf{Name}               & \textbf{Description} & \textbf{Group} & \textbf{N}           & \textbf{E}              & \textbf{C} & \textbf{Dlog} \\
\hline
LeGresley\_4908     & Power Network                   & LaGresley                                         & 4908   & 17984   & 0.76 & -2.83 \\
cavity05            & Fluid dynamics                  & DRIVCAV                                           & 1182   & 18330   & 0.77 & -1.58 \\
crystm01            & 3D Model                        & Boeing                                            & 1625   & 18369   & 0.53 & -1.86 \\
msc01440            & 3D Model                        & Boeing                                            & 1440   & 23855   & 0.51 & -1.64 \\
bcsstm27            & 3D Model                        & HB & 1224   & 28675   & 0.67 & -1.42 \\
dictionary28        & Word asociation                 & Pajek                                             & 24831  & 71014   & 0.24 & -3.64 \\
ca-CondMat          & Collaboration                   & SANP                                              & 21363  & 91342   & 0.65 & -3.40 \\
foldoc              & Word asociation                 & Pajek                                             & 13356  & 91471   & 0.33 & -2.99 \\
wiki-Vote           & Social                          & SNAP                                              & 7066   & 100736  & 0.15 & -2.39 \\
ca-HepPh            & Collaboration                   & SANP                                              & 11204  & 117649  & 0.59 & -2.73 \\
Wordnet3            & Word asociation                 & Pajek                                             & 75606  & 120472  & 0.04 & -4.38 \\
p2p-Gnutella31      & Social                          & SNAP                                              & 62561  & 147878  & 0.00 & -4.12 \\
lhr07c              & Chimestry                       & Mallya                                            & 7337   & 155150  & 0.02 & -2.24 \\
Na5                 & Chimestry                       & PARSEC                                            & 5832   & 155731  & 0.39 & -2.04 \\
usroads-48          & Roads                           & Gleich                                            & 126146 & 161950  & 0.02 & -4.69 \\
email-Enron         & Email                           & SNAP                                              & 33696  & 180811  & 0.49 & -3.50 \\
ca-AstroPh          & Collaboration                   & SANP                                              & 17903  & 197031  & 0.66 & -2.91 \\
TSOPF\_RS\_b162\_c1 & Electric Circuit                & TSOPF                                             & 5374   & 202415  & 0.09 & -1.85 \\
internet            & Computer Networks               & Pajek                                             & 124651 & 205805  & 0.10 & -4.58 \\
qc2534              & Electromagnetic                 & Bai                                               & 2534   & 232947  & 0.99 & -1.14 \\
cage11              & DNA                             & vanHeukelum                                       & 39082  & 299402  & 0.28 & -3.41 \\
EAT\_SR             & Word asociation                 & Pajek                                             & 23218  & 305498  & 0.10 & -2.95 \\
nemeth17            & Chimestry                       & Nemeth                                            & 9506   & 319563  & 0.74 & -2.15 \\
email-EuAll         & Email                           & SNAP                                              & 224832 & 340795  & 0.06 & -4.87 \\
rajat17             & Electric Circuit                & Rajat                                             & 93342  & 367910  & 0.73 & -4.07 \\
soc-Epinions1       & Social                          & SNAP                                              & 75877  & 405739  & 0.13 & -3.85 \\
psmigr\_3           & Migrations                      & HB                                                & 3140   & 413921  & 0.48 & -1.08 \\
TSC\_OPF\_300       & Electric Circuit                & IPSO                                              & 9773   & 415288  & 0.36 & -2.06 \\
cit-HepPh           & Citation                        & SNAP                                              & 34401  & 420828  & 0.29 & -3.15 \\
soc-Slashdot0811    & Social                          & SNAP                                              & 77360  & 546487  & 0.62 & -3.74 \\
patents\_main       & Citations                       & Pajek                                             & 230686 & 554949  & 0.04 & -4.68 \\
fe\_rotor           & Fisics                          & DINAMICS10                                        & 99617  & 662431  & 0.40 & -3.87 \\
dblp-2010           & Collaboration                   & LAW                                               & 226413 & 716460  & 0.64 & -4.55 \\
598a                & Fisics                          & DINAMICS10                                        & 110971 & 741934  & 0.43 & -3.92 \\
sx-superuser        & Social                          & SNAP                                              & 189191 & 781375  & 0.26 & -4.36 \\
coAuthorsCiteseer   & Collaboration                   & DIMACS10                                          & 227320 & 814134  & 0.69 & -4.50 \\
oh2010              & Borders & DIMACS10                                          & 365344 & 884120  & 0.38 & -4.88 \\
com-Amazon          & Product co-purchasing           & SNAP                                              & 334863 & 925872  & 0.40 & -4.78 \\
loc-Gowalla         & Social                          & SNAP                                              & 196591 & 950327  & 0.24 & -4.31 \\
web-NotreDame       & Web                             & SNAP                                              & 325729 & 1117563 & 0.24 & -4.68 \\
language            & NLP                             & Tromble                                           & 399130 & 1192675 & 0.56 & -4.82 \\
Linux\_call\_graph  & Call graph                      & Sorensen                                          & 317926 & 1207269 & 0.09 & -4.62
\end{tabular}
\end{table}

\end{document}